\documentclass{article}

\usepackage{microtype}
\usepackage{graphicx}
\usepackage{booktabs} 
\usepackage{multirow}
\usepackage{float}
\usepackage[utf8]{inputenc} 
\usepackage[T1]{fontenc}    
\usepackage{url}            
\usepackage{booktabs}       
\usepackage{amsfonts}       
\usepackage{nicefrac}       
\usepackage{microtype}      
\usepackage{mathtools}
\usepackage{natbib}
\usepackage{xcolor}
\usepackage{url}
\usepackage{amsmath}
\usepackage{amssymb}
\usepackage{amsthm}
\usepackage{graphicx}
\usepackage{nicefrac}
\usepackage{appendix}
\usepackage{enumitem}
\usepackage{xspace}
\usepackage{wrapfig}
\usepackage{caption}
\usepackage{subcaption}
\usepackage{array}
\definecolor{darker}{rgb}{0,0.15,0.7}
\usepackage[colorlinks, urlcolor=darker, citecolor=darker, linkcolor=darker]{hyperref} 
\setcitestyle{round}

\newcommand{\method}{{InstructRetro}\xspace}
\newcommand{\retro}{{Retro}\xspace}
\newcommand{\instructgpt}{\textcolor{black}{Instruct$\text{GPT}_\text{RAG}$}\xspace}
\usepackage{hyperref}

\usepackage[accepted]{icml2024}

\usepackage{amsmath}
\usepackage{amssymb}
\usepackage{mathtools}
\usepackage{amsthm}

\usepackage[capitalize,noabbrev]{cleveref}
\usepackage{svg}
\usepackage{scalerel,graphicx,xparse}
\NewDocumentCommand\emojiseed{}{{\includegraphics[scale=0.4]{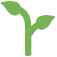}}}
\NewDocumentCommand\emojiherb{}{{\includegraphics[scale=0.4]{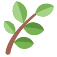}}}
\theoremstyle{plain}

\theoremstyle{definition}

\theoremstyle{remark}

\usepackage{soul}
\usepackage[textsize=tiny]{todonotes}

\icmltitlerunning{InstructRetro: Instruction Tuning post Retrieval-Augmented Pretraining}

\begin{document}

\twocolumn[
\icmltitle{InstructRetro: Instruction Tuning post  Retrieval-Augmented Pretraining}




\begin{icmlauthorlist}
\icmlauthor{Boxin Wang}{comp}
\icmlauthor{Wei Ping}{comp}
\icmlauthor{Lawrence McAfee}{comp}
\icmlauthor{Peng Xu}{comp}
\icmlauthor{Bo Li}{sch}
\icmlauthor{Mohammad Shoeybi}{comp}
\icmlauthor{Bryan Catanzaro}{comp}
\end{icmlauthorlist}

\icmlaffiliation{comp}{NVIDIA}
\icmlaffiliation{sch}{UIUC}

\icmlcorrespondingauthor{Boxin Wang\\}{boxinw@nvidia.com}
\icmlcorrespondingauthor{~Wei Ping}{wping@nvidia.com}

\icmlkeywords{Machine Learning, ICML}

\vskip 0.3in
]



\printAffiliationsAndNotice{}  

\begin{abstract}
Pretraining auto-regressive large language models~(LLMs) with retrieval demonstrates better perplexity and factual accuracy by leveraging external databases. 
However, the size of existing pretrained retrieval-augmented LLM is still limited~(e.g., \emph{Retro} has 7.5B parameters), which limits the effectiveness of instruction tuning and zero-shot generalization.
In this work, we introduce \emph{Retro} 48B, the largest LLM pretrained with retrieval. 
Specifically, we continue to pretrain a 43B GPT model on additional 100 billion tokens using the {Retro} augmentation method by retrieving from 1.2 trillion tokens.
Notably, the obtained foundation model, Retro 48B, largely outperforms the counterpart GPT 43B trained on 1.2T tokens in terms of perplexity with only 2.58\% additional GPU hours, demonstrating the significant scaling potential of the method.
After instruction tuning on \retro, \emph{InstructRetro} demonstrates significant improvement over the instruction tuned GPT on a wide range of zero-shot tasks. 
Specifically, the average improvement of \method is 7\% over its GPT counterpart across 8 short-form QA and reading comprehension tasks, 10\% over GPT across 4 challenging long-form QA tasks, and 16\% over GPT across 3 summarization tasks. 
Surprisingly, we find that one can ablate the encoder from \method architecture and directly use its decoder backbone, while achieving comparable results. 
Our results highlight the promising direction to obtain a better GPT decoder through continued pretraining with retrieval before instruction tuning.
Our code and checkpoints are publicly available at: \url{https://huggingface.co/nvidia/retro-48b-instruct-4k}.
\end{abstract}

\section{Introduction}
\label{sec:intro}

Retrieval helps large language models~(LLM) to handle current events, detailed knowledge, proprietary information not in pretraining, and to improve factual grounding~\citep[e.g.,][]{nakano2021webgpt, thoppilan2022lamda, borgeaud2022improving}. 
In the previous study, pretraining auto-regressive language model with retrieval~(i.e., \emph{Retro}) demonstrates successes in reducing perplexity~\citep{borgeaud2022improving} and improving factual accuracy~\citep{wang2023shall}.

In the past year, the decoder-only auto-regressive LLMs have demonstrated remarkable successes~\citep[e.g.,][]{chatgpt, openai2023gpt4}, because 
\emph{i)} LLMs have been scaled to hundreds of billion parameters~\citep{brown2020language, rae2021scaling, smith2022using, chowdhery2022palm},
\emph{ii)} pretraining corpus has been scaled up to trillions of tokens~\citep{hoffmann2022training, llama, llama2},
and \emph{iii)}  instruction tuning~\citep{wei2021finetuned,chung2022scaling} and reinforcement learning from human feedback~(RLHF)~\citep{ouyang2022training} recipes have been applied on these pretrained LLMs.

In contrast, the pretrained retrieval-augmented language models still have a relatively small number of parameters trained with a limited number of tokens. 
For example, the auto-regressive \emph{Retro} has 7.5B parameters and is trained on 600B tokens~\citep{borgeaud2022improving}, \emph{Retro++} has 9.5B parameters and is trained on 330B tokens~\citep{wang2023shall}, and T5-based \emph{Atlas} has 11B parameters and is trained with retrieval on maximum 327M tokens~\citep{izacard2022few}.
In addition, none of previous models have been applied with instruction tuning and RLHF to enhance usability.
The lack of scaling could also limit the effectiveness of instruction tuning~\citep{wei2021finetuned} and other intriguing properties that exist in large language models~\citep{wei2022emergent}.

In this work, we scale up \emph{Retro} up to 48B parameters, trained on 1.2T tokens in total, i.e., 1.1T tokens for pretraining its GPT backbone, 100B tokens for continued retrieval-augmented pretraining while retrieving from 1.2T tokens.
As a result, we can mitigate the zero-shot generalization gap on  {a wide range of tasks} after applying instruction tuning.

\begin{figure*}[t]
    \centering
    \includegraphics[width=0.8\linewidth]{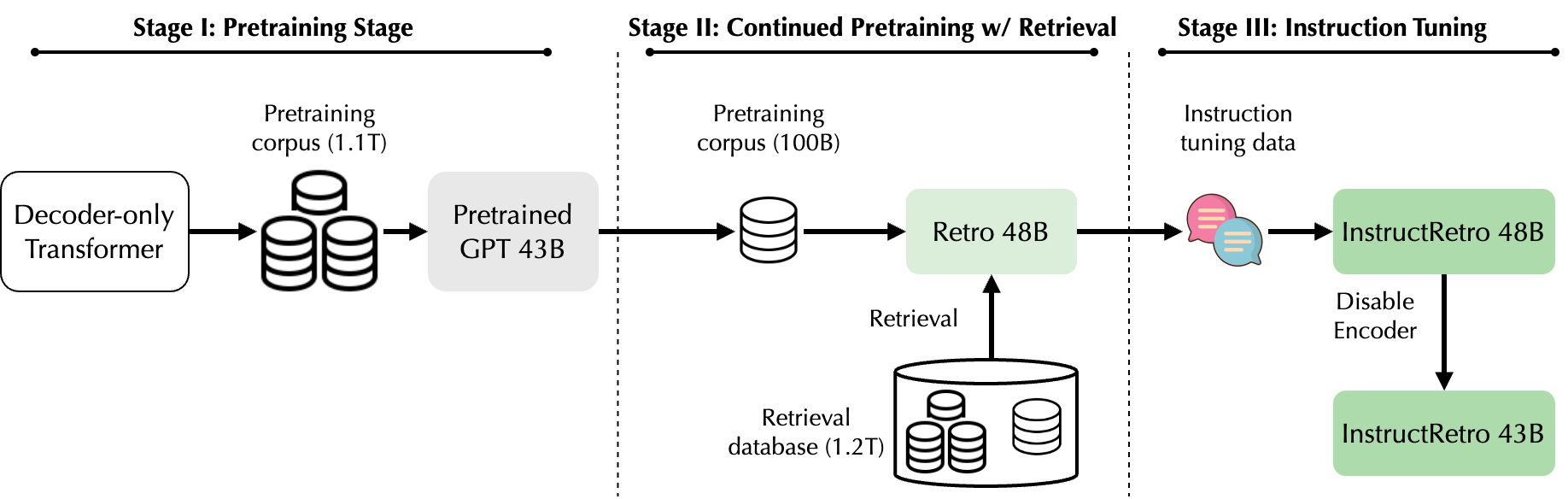}
    \caption{Training pipeline for \method 48B and \method 43B.}
    \label{fig:pipeline}
\end{figure*}

Specifically, we make the following contributions: 
\vspace{-0.1cm}
\begin{enumerate}[leftmargin=3.1em]
    \vspace{-0.2cm}
    \item We introduce \emph{Retro} 48B, the largest LLM pretrained with retrieval. 
    To save the computation budget,  we continue to pretrain a 43B parameter GPT model~(originally trained on 1.1T tokens) on additional 100B tokens by retrieving from 1.2T tokens. 
    In contrast to \emph{Retro-fitting}~\citep{borgeaud2022improving}, that freezes pretrained decoder weights, we unfreeze the decoder, jointly train all the parameters and find better perplexity.~\footnote{Note that, it turns out that unfreezing of decoder is an important design not only for better perplexity, and it eventually leads to the interesting finding after instruction tuning.}
    Notably, with only 2.58\% additional GPU hours, the perplexity improvement of Retro 48B over its GPT 43B counterpart is still significant even at this scale, demonstrating that the value of retrieval does not diminish with scaling model size.
    \vspace{-0.2cm}
    \item  After instruction tuning, \emph{InstructRetro} 48B demonstrates strong zero-shot capability to incorporate context for {various downstream tasks}, and significantly outperforms instruction-tuned GPT with retrieval-augmented generation (RAG). The training pipeline of \method is shown in Figure \ref{fig:pipeline}.
    \vspace{-0.2cm}
    \item Perhaps surprisingly, we find that one can directly ablate the encoder from \emph{IntructRetro} 48B. The obtained decoder-only \emph{IntructRetro} 43B still achieves very comparable results on downstream tasks. This highlights the promising direction of obtaining better decoder-only LLMs through continued pretraining with retrieval before instruction tuning.
\vspace{-0.3cm}
\end{enumerate}

We organize the rest of the paper as follows.
We discuss related work in \S~\ref{sec:related_work}. 
We introduce the continued pretraining of {Retro 48B} in \S~\ref{sec:pretraining} and the instruction tuning recipe in \S~\ref{sec:instruction}.
We report results in Section~\ref{sec:experiment} and conclude the paper in \S~\ref{sec:conclusion}.

\section{Related Work}
\label{sec:related_work}

%
Retrieval-augmented language models have been established for open domain question answering for years~\citep{karpukhin2020dense, lewis2020retrieval, guu2020retrieval, borgeaud2022improving, izacard2022few}.
In the previous study, language models have been augmented with retrieval at inference~\citep{khandelwal2019generalization,yogatama2021adaptive}, fine-tuning~\citep{karpukhin2020dense, lewis2020retrieval, guu2020retrieval, huang2023raven, shi2023replug}, and pretraining~\citep{borgeaud2022improving, izacard2022few, wang2023shall,shi2023context}.
Retrieval-augmented pretraining is particularly interesting, as it can largely reduce model perplexity~\citep{borgeaud2022improving}, enhance factuality~\citep{wang2023shall}, and improve downstream task accuracy after task-specific fine-tuning~\citep{izacard2022few} and reasoning capability \citep{shi2023context}.

In contrast to the state-of-the-art decoder-only LLMs with hundreds of billion parameters~\citep{gpt3, rae2021scaling, smith2022using, chowdhery2022palm}, the sizes of pretrained retrieval-augmented LLMs are still around 10B parameters~\citep{borgeaud2022improving, wang2023shall, izacard2022atlas}, which largely limits the zero-shot generalization capability after instruction tuning~\citep{wei2021finetuned, ouyang2022training, chung2022scaling}. 
For example, \citet{wei2021finetuned} find instruction tuning to be more effective when the decoder-only LLM has around 50B parameters.

%
Instruction tuning aims to teach LLMs to follow natural language instructions~\citep{wei2021finetuned, ouyang2022training, sanh2021multitask, mishra2021cross}, which becomes an indispensable ingredient to build the state-of-the-art LLMs for downstream tasks~\citep{chatgpt, openai2023gpt4, llama2}.
In the past years, many high-quality instruction tuning datasets have been created, including FLAN~\citep{chung2022scaling}, OpenAssistant \citep{köpf2023openassistant},  Self-Instruct~\cite{wang2022self}, Dolly~\citep{DatabricksBlog2023DollyV2}, Unnatural Instructions~\cite{honovich2022unnatural}.
{A concurrent work, RA-DIT \citep{lin2024radit}, focuses on retrieval-augmented instruction tuning and further augments 20 instruction tuning datasets with retrieval, which supports fine-tuning both LLM and retriever to yield high-quality neighbors.
In contrast, our work focuses on retrieval-augmented pretraining, which extends the scale of the retrieval database to trillions of tokens. 
Although the two work are orthogonal, \method 43B outperforms RA-DIT 65B on certain benchmarks as shown in Table~\ref{tab:shortqa}.
We leave it as an interesting future direction to apply RA-DIT retrieval-augmented instruction tuning data to the instruction tuning stage of \method for further performance improvement of retrieval-augmented LLMs.
}

\section{Continued Pretraining of GPT with Retrieval}
\label{sec:pretraining}
In this section, we start by introducing the preliminaries of \emph{\retro} \citep{borgeaud2022improving} and highlight some key differences between \retro and GPT. 
We then go through the pretraining details of how we scale up the size of \retro to 48B, a size that has never been studied before.

\subsection{Preliminaries of Retro}

\retro \citep{borgeaud2022improving} is an auto-regressive language model pretrained with retrieval augmentation.
While \retro shares the backbone of GPT models, \retro differs from GPT by incorporating an additional \textit{Retro encoder}.
The Retro encoder is adept at encoding features of retrieved neighbors from \textit{external knowledge bases}. 
Furthermore, \retro adds \textit{chunk-wise cross-attention} layers within its decoder transformer architecture to integrate retrieved information from the Retro encoder effectively.
This design paradigm also makes \retro different from the encoder-decoder architecture (e.g., T5 \citep{raffel2020exploring} and Atlas \citep{izacard2022atlas}).
The success of scaling decoder-only autoregressive language models (e.g., ChatGPT \citep{chatgpt} and GPT-4 \citep{openai2023gpt4}) motivates us to further scale up autoregressive \retro and understand the potential benifit of retrieval-augmented pretraining.

\textbf{Retro encoder} is a shallow bidirectional transformer to encode retrieved neighbors from external databases into dense features. Specifically, in this work, we follow \citet{borgeaud2022improving} and use a two-layer bidirectional transformer as the \retro encoder with the same hidden dimension as the \retro backbone decoder. 
Our preliminary results show that increasing the layers of the \retro encoder does not bring better perplexity on the validation set, but only increases the computational overhead and model parameters. 

\textbf{Retrieval database.} 
\citet{borgeaud2022improving} demonstrates that retrieval-augmented pretraining can significantly benefit from large-scale retrieval up to trillions of tokens.
To build the retrieval database, we utilize the entire pretraining corpus, but holding out $1\%$ as a validation set. 
This ensures that both \retro and GPT models are pretrained on an equivalent volume of information from the pretraining corpus. 
Our retrieval database is a key-value database, where values are chunks of tokens split from the pretraining corpus, and the keys are corresponding BERT embeddings \citep{devlin2018bert}. 
The pretraining corpus consists of 1.2 trillion tokens of English corpus. More details of the pretraining corpus can be found in Appendix \S\ref{app:corpus}.
In summary, our retrieval database comprises 19 billion chunks, with each chunk containing 64 tokens.

\textbf{Chunk-wise cross-attention.} Aligning with the chunk-wise design of the retrieval database, \retro splits the input tokens into a sequence of chunks. Specifically, \retro retrieves nearest neighbor chunks using the previous chunk and fuses this information with the context from preceding chunks to guide the generation of the next chunk. Formally, given a input sequence $X$ with $n$ tokens $X=(x_1, ..., x_n)$, \retro splits $X$ into a sequence of $l$ chunks $(C_1,...,C_l)$ with chunk size $m = \frac{n}{l}$. 
From a high-level perspective, \retro uses the last $(i-1)$-th chunk $C_{i-1}$ to retrieve $k$ nearest neighbor chunks $\mathcal{N}(C_{i-1})$ from the retrieval database, and fuses the contextual information from the previous chunks $(C_1,...,C_{i-1})$ and retrieval information from $\mathcal{N}(C_{i-1})$ by cross-attention to guide the generation of the next $(i)$-th chunk $C_{i}$.
To avoid breaking the causality, the autoregressive generation of $i$-th chunk $C_i$ can only use the nearest neighbors of the previous chunk $\mathcal{N}(C_{i-1})$ instead of $\mathcal{N}(C_{i})$. 
In our work, we follow \citet{borgeaud2022improving} and retrieve top-$k=2$ nearest neighbors for each chunk, with chunk size $m=64$ and the maximum number of tokens $n=4096$.

\subsection{Retro-fitting: continued pretraining with retrieval}
There are two main challenges of scaling up \retro: the large-scale retrieval database and the pretraining cost of LLMs. 
To overcome the challenges,
we leverage the \emph{Faiss} index \citep{faiss} to achieve fast approximate nearest neighbor search and \emph{retro-fitting} technique to reuse the pretrained GPT parameters and save computational cost.

\textbf{Retrieval index to the large-scale retrieval database.}
We use the Faiss index \citep{faiss} as the implementation for the dense retriever to search for approximate nearest neighbors in the BERT embedding space.
We configure the Faiss index to cluster the dense embeddings into $2^{22}$ centroids accelerated with Hierarchical Navigable Small World (HNSW) graphs \citep{hnsw} to speed up the query. 
We also encode the embeddings with optimized product quantization \citep{pq,opq} to compress memory overhead and further improve the query throughput.
As a result, we can achieve 4\textit{ms} per query over the whole pretraining corpus averaged for each chunk on a DGX-A100 node.
One may find more details in Appendix~\S\ref{app:faiss}.


\textbf{Base pretrained GPT.}
We launch continued pretraining (\textit{i.e.}, GPT-fitting and Retro-fitting) based on pretrained GPT models.
Specifically, we pretrain from scratch a set of GPT models with the following parameter sizes: 823M, 2.25B, 8.5B, 22B, and 43B. 
All of the models are based on Transformer~\citep{transformers} with different hidden dimensions, number of layers, and attention heads. 
We adopt the Sentence Piece tokenizer \citep{kudo2018sentencepiece} for both GPT and \retro.
We pretrain all models with 1.1 trillion tokens of the pretraining corpus.
More details of corpus be found in Appendix \S\ref{app:corpus}.

\begin{figure*}[t]
    \centering
    \includegraphics[width=0.8\linewidth]{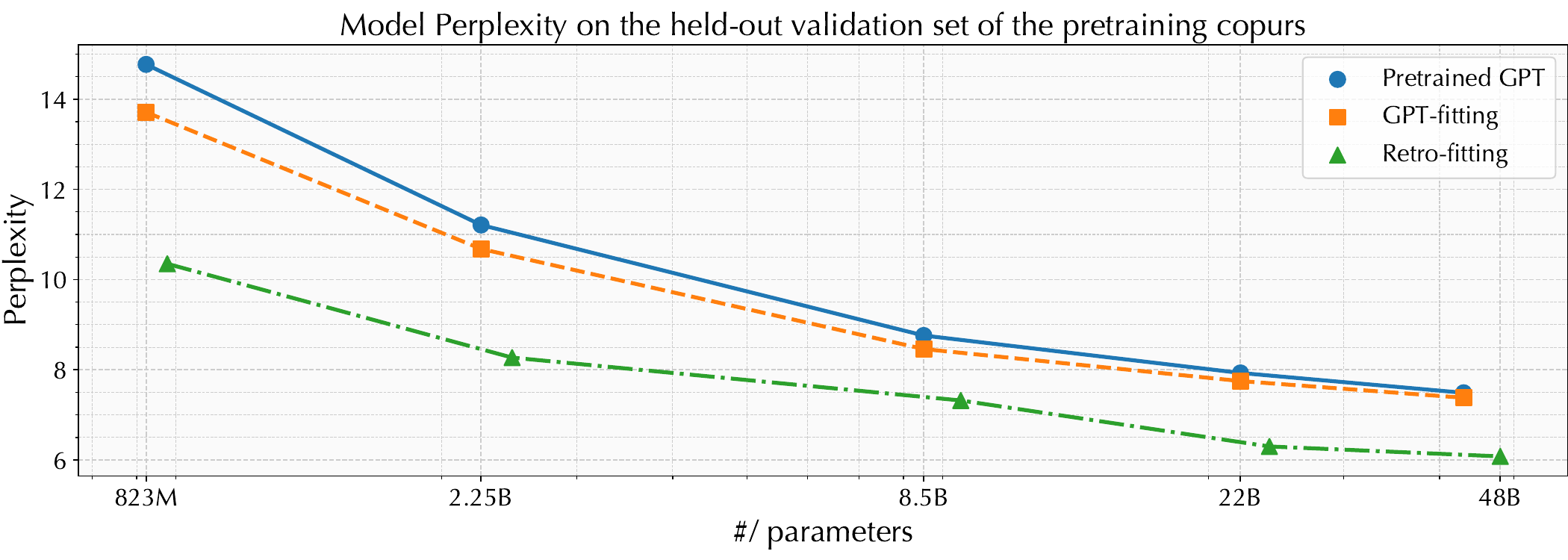}
    \vspace{-1em}
    \caption{Perplexity evaluation of pretrained GPT models, GPT-fitting, and Retro-fitting models across various parameter sizes on the held-out validation set. In contrast to~\citet{borgeaud2022improving}, we unfreeze all parameters for Retro-fitting.  \retro significantly outperforms GPT models, achieving the perplexity comparable to GPT models with 4$\times$ larger parameter sizes.}
    \label{fig:ppl}
\end{figure*}

\begin{figure}[htp!]
  \begin{center}
    \includegraphics[width=0.8\linewidth]{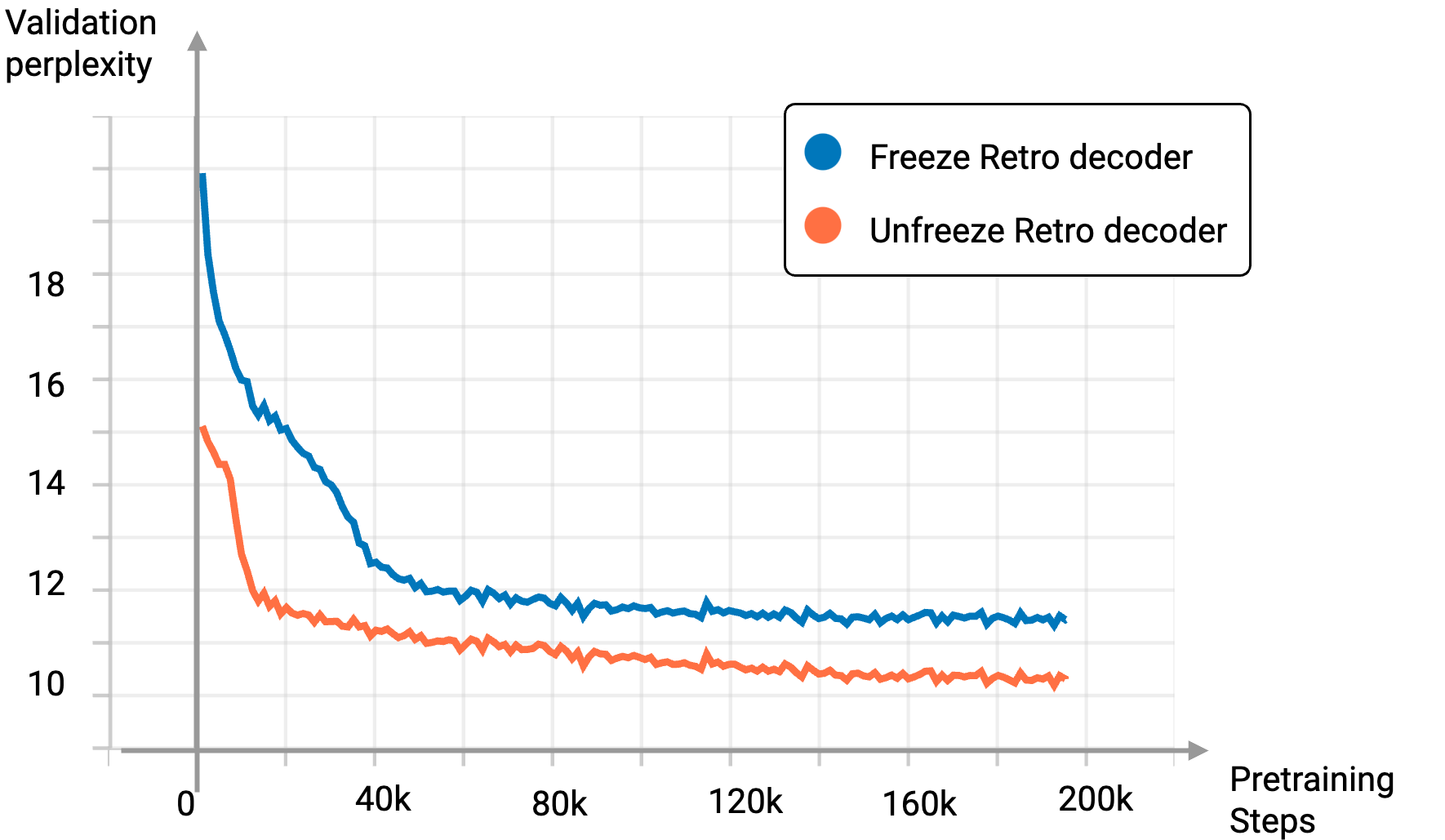}
  \end{center}
  \vspace{-3mm}
  \caption{Validation perplexity of Retro-fitting (823M) when we freeze or unfreeze \retro decoder during continued pretraining on 100B tokens.}    \label{fig:decoder}
\end{figure}

\textbf{Unfreezing decoder at Retro-fitting.} As \retro shares its backbone decoder with the GPT decoder and only adds around 10\% additional parameters for \retro encoder and cross-attention, we can initialize \retro decoder from  pretrained GPT models, randomly initialize \retro encoder and cross-attention, 
and continue pretraining with retrieval, which is named as ``\textit{Retro-fitting}''. 
Note that, \citet{borgeaud2022improving} freezes the decoder parameters at {Retro-fitting}.
{In contrast, we \textbf{unfreeze all the decoder parameters} and continue pretraining the entire model.}
We also conduct an ablation study of \retro-fitting based on a pretraiend GPT of 823M parameters and compare the validation perplexity loss when freezing or unfreezing \retro decoder during pretraining.
As shown in Figure \ref{fig:decoder}, given the same training schedules, unfreezing \retro decoder parameters converges faster and demonstrates better validation perplexity, which eventually yields a better \retro decoder to incorporate in-context retrieved evidence, even without a \retro encoder as shown in \S\ref{sec:ablation}.
We continue pretraining with retrieval on an additional 100 billion tokens, which is 9\% of the pretraining data used for pretrained GPT models.
To have a fair comparison, we also continue pretraining GPT foundation models on the same 100 billion tokens, which we name ``\textit{GPT-fitting}''. 
In terms of overall pretraining cost, \retro 48B only need 2.58\% additional GPU hours than its counterpart GPT trained on 1.2T tokens.
More details of continued pretraining are in Appendix \ref{app:schedule} and \ref{app:computational-cost}.

\textbf{Perplexity evaluation.}
We evaluate the perplexity of GPT foundation models, GPT-fitting models, and \retro-fitting models of varying parameter sizes in Figure \ref{fig:ppl}. 
The validation corpus consists of 1\%  held-out samples from the pretraining corpus, which are not used in the pretraining stage, the continued pretraining stage, and the retrieval database to ensure that there is no validation data leakage.
From Figure~\ref{fig:ppl}, one can see that after continued pretraining on additional 100 billion tokens, the perplexity of GPT-fitting slightly improves over original pretrained GPT, while \retro significantly outperforms both GPT and GPT-fitting across different parameter sizes in terms of perplexity. 
\retro achieves even better perplexity than GPT models with 4$\times$ larger parameter sizes. Notably the improvement is still significant when the parameter sizes of \retro scale up to 48B, and the gap does not decrease from 8B to 48B.
We present more evaluation results in \S\ref{sec:ablation}.

\section{Instruction Tuning}
\label{sec:instruction}

Instruction tuning can significantly improve the ability of foundation LLMs to follow instructions, thus improving zero-shot results on downstream tasks~\citep[e.g.,][]{wei2021finetuned, chung2022scaling}. 
In this section, we further enhance \retro via instruction tuning. 

\begin{figure*}[t]
    \centering
    \includegraphics[width=\linewidth]{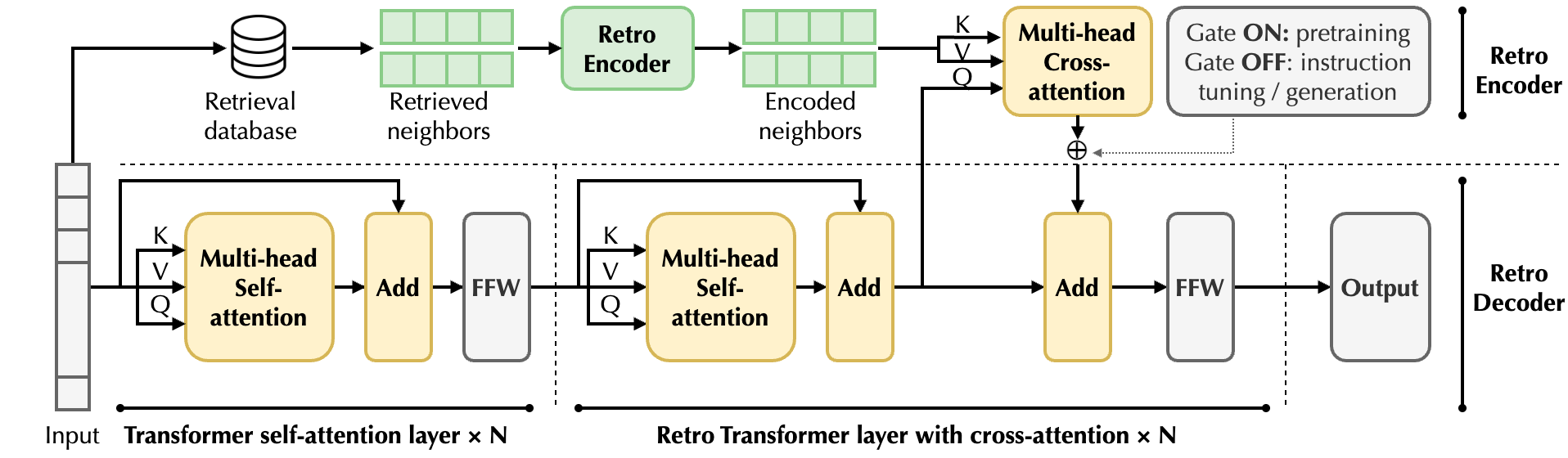}
    \caption{ 
    Simplified architecture diagram of \method. We omit the layer norm, softmax, and embedding layers for simplicity. We add additional 0/1 gates between cross-attention output and the residual connection from self-attention output. 
    During pretraining, we keep the \retro encoder gate ON with gate-value 1. During instruction tuning and inference, we bypass the \retro encoder by turning the \retro encoder gate OFF with gate-value 0 to solely serve as a GPT decoder. 
    }
    \label{fig:retro}
\end{figure*}

\subsection{Datasets Blending}
Existing instruction tuning methods mainly leverage supervised fine-tuning on a blend of instruction following datasets \citep{wei2021finetuned, chung2022scaling,sanh2022multitask,wang2023far}.

We use a blend of high-quality instruction tuning datasets to train LLMs to follow instructions in conversational formats, which include: \emph{i)} a high-quality social dialogue dataset SODA \citep{kim2022soda}, \emph{ii)} a long-form QA dataset ELI5 that requires elaborate answers \citep{fan2019eli5}, \emph{iii)} LLM-generated instructions: Self-Instruct \citep{wang2022selfinstruct}
 and Unnatural Instructions \citep{honovich2022unnatural}, \emph{iv)} FLAN and Chain-of-thought datasets \citep{chung2022scaling,wei2022chain,longpre2023flan}, \emph{v}) a private crowd-sourced conversational dataset and public human-written conversation datasets OpenAssistant \citep{köpf2023openassistant} and Dolly \citep{DatabricksBlog2023DollyV2}, and \emph{vi}) samples from the pretraining corpus. 
 
 The format of all the instruction tuning data is unified in a conversational way with three roles: ``system'', ``assistant'', and  ``user''. The ``system'' role sets up the tone and style of LLM assistants to give helpful, detailed, and polite answers to the user's questions. The ``user'' and ``assistant'' role contains the questions and the corresponding answers from the instruction tuning datasets. We show an example format of the instruction data in Appendix \ref{app:example}. 
 In total, we collect a total of 128K high-quality samples for instruction tuning.

\subsection{Training details}

For each training sample, we take the multi-turn conversations between the user and the assistant as context and apply the loss mask only to the last response from the assistant.
We use the standard language modeling loss with teacher forcing. 
Since \citet{wei2021finetuned} suggests that instruction tuning is most effective with \textit{large} language models, we apply instruction tuning to the GPT-fitting 43B model and the \retro 48B model, naming them ``\instructgpt 43B''\footnote{We distinguish ``\instructgpt'', which uses supervised fine-tuning and RAG, from ``InstructGPT'' \citep{ouyang2022training}, which leverage RLHF for instructing tuning.} and ``\method 48B'', respectively. 
We finetune the LLMs by taking the loss only on the answer part with a batch size of 128 and a learning rate of 5e-6 for 1000 steps with a weight decay of 0.01. 
We use the Adam optimizer \citep{adam} with $\beta_1=0.9$ and $\beta_2=0.98$.

\textbf{Instruction tuning for \retro.} 
Since the \retro backbone largely shares with GPT models, the training objective of \retro is also the same as GPT models. 
However, one noticeable difference is that \retro requires retrieval of nearest neighbors, which is not available from all the instruction tuning datasets. 
Since the instruction tuning data is high-quality, retrieval from the pretraining corpus can yield noisy neighbors, thus not helping improve the model capabilities to follow instructions.
We instead disable the \retro encoder by skipping the cross-attention connection through a manually-set gated mechanism as detailed in Figure \ref{fig:retro}, which sets the gate to \emph{zero} when retrieved neighbors are not available. 
During backpropagation, as the cross-attention module and the connected retro encoder are skipped, their parameters are effectively frozen, and only the weights of the decoder backbone get updated.
Such design not only simplifies the instruction tuning and inference but also makes \retro learn to inference with and without retrieval during instruction tuning, potentially improving the generalization of the \retro decoder.
We also leave it as an important future direction to construct retrieval-augmented instruction tuning data for retrieval-augmented generation.


\begin{table*}[t]\small
\centering
\caption{Zero-shot evaluation on {eight}  short-form QA and reading comprehension datasets. The average \textit{relative} improvement of decoder-only \method 43B across the short-form QA tasks is 7\% over \instructgpt.
Note that, \method 43B obtains very comparable results than  original \method 48B with encoder.
\emojiherb denotes using retrieval augmentation at both training and generation, while 
\emojiseed denotes using retrieval augmentation at inference only.
}
\label{tab:shortqa}
\vspace{5pt}
\begin{tabular}{l|cccccccc}
\toprule
Task & NQ & TriviaQA & NewsQA & SQuAD 2.0  & SQuAD 1.1  & Quoref & NarrativeQA & DROP \\
\midrule
Metric & EM & EM & F1 & F1 / EM & F1 / EM & F1 & F1 & F1 \\
\midrule
GPT-3 175B   & \multirow{2}{*}{14.6} & \multirow{2}{*}{64.3} & \multirow{2}{*}{-} & \multirow{2}{*}{59.5 / 52.6} & \multirow{2}{*}{-} & \multirow{2}{*}{-} & \multirow{2}{*}{-} & \multirow{2}{*}{23.6} \\
(\citeauthor{gpt3}) \\ 
\midrule
PaLM 2 -L   & \multirow{2}{*}{37.5} & \multirow{2}{*}{-} & \multirow{2}{*}{-} & \multirow{2}{*}{- / -} & \multirow{2}{*}{-} & \multirow{2}{*}{-} & \multirow{2}{*}{-} & \multirow{2}{*}{-} \\
(\citeauthor{chowdhery2022palm}) \\
\midrule
GLaM 64B   & \multirow{2}{*}{24.7} & \multirow{2}{*}{{71.3}} & \multirow{2}{*}{-} & \multirow{2}{*}{71.1 / 64.7} & \multirow{2}{*}{- / -} & \multirow{2}{*}{-} & \multirow{2}{*}{-} & \multirow{2}{*}{\textbf{57.3}} \\
(\citeauthor{du2021glam}) \\
\midrule
FLAN-LaMDA 137B  & \multirow{2}{*}{20.7} & \multirow{2}{*}{68.1} & \multirow{2}{*}{-} & \multirow{2}{*}{44.2 / -} & \multirow{2}{*}{\textbf{80.1} / -} & \multirow{2}{*}{-} & \multirow{2}{*}{-} & \multirow{2}{*}{22.7} \\
(\citeauthor{wei2021finetuned}) \\
\midrule
Llama 2 RAG 70B \hfill \emojiseed& \multirow{2}{*}{37.7} & \multirow{2}{*}{65.6} & \multirow{2}{*}{53.4} & \multirow{2}{*}{71.4 / 64.1} & \multirow{2}{*}{73.4 / 66.2} & \multirow{2}{*}{69.7} & \multirow{2}{*}{52.7} & \multirow{2}{*}{57.2} \\
(\citeauthor{llama2}) \\
\midrule
Retro 7.5B \hfill  \emojiherb& \multirow{2}{*}{8.9} & \multirow{2}{*}{36.0} & \multirow{2}{*}{-} & \multirow{2}{*}{- / -} & \multirow{2}{*}{-} & \multirow{2}{*}{-} & \multirow{2}{*}{-} & \multirow{2}{*}{-} \\
(\citeauthor{borgeaud2022improving}) \\
\midrule
Retro++ 9B  \hfill \emojiherb& \multirow{2}{*}{25.8} & \multirow{2}{*}{48.3} & \multirow{2}{*}{-} & \multirow{2}{*}{- / -} & \multirow{2}{*}{-} & \multirow{2}{*}{-} & \multirow{2}{*}{-} & \multirow{2}{*}{-} \\
(\citeauthor{wang2023shall}) \\
\midrule
Atlas 11B  \hfill \emojiherb& \multirow{2}{*}{26.7} & \multirow{2}{*}{56.9} & \multirow{2}{*}{-} & \multirow{2}{*}{- / -} & \multirow{2}{*}{- / -} & \multirow{2}{*}{-} & \multirow{2}{*}{-} & \multirow{2}{*}{-} \\
(\citeauthor{izacard2022atlas}) \\
\midrule
Raven 11B  \hfill \emojiherb& \multirow{2}{*}{29.6} & \multirow{2}{*}{65.7} & \multirow{2}{*}{-} & \multirow{2}{*}{- / -} & \multirow{2}{*}{- / -} & \multirow{2}{*}{-} & \multirow{2}{*}{-} & \multirow{2}{*}{-} \\
(\citeauthor{huang2023raven}) \\
\midrule
RA-DIT 65B \hfill \emojiherb & \multirow{2}{*}{35.2} & \multirow{2}{*}{75.4} &  \multirow{2}{*}{-} & \multirow{2}{*}{-} & \multirow{2}{*}{- / -} & \multirow{2}{*}{-} & \multirow{2}{*}{-} & \multirow{2}{*}{-} \\
(\citeauthor{lin2024radit}) \\
\midrule
\instructgpt 43B  \hfill \emojiseed& 37.0 & 78.1 &  52.4 & 70.7 / 64.3 & 72.4 / 65.8  & 71.5 & 53.9 & 51.8  \vspace{.15cm}\\
InstructRetro 43B \hfill  \emojiherb& \textbf{38.9} & \textbf{78.3} & \textbf{57.4} & \textbf{75.6} / \textbf{69.3} & 77.1 / 70.4 & \textbf{76.2} & \textbf{60.0} & {54.8} \\
\tiny{(w/o encoder, Avg: +7\%)} & \tiny{(+5.14\%)} & \tiny{(+0.26\%)} & \tiny{(+9.54\%)} & \tiny{(+6.93\%)} & \tiny{(+6.49\%)} & \tiny{(+6.57\%)} & \tiny{(+11.32\%)} & \tiny{(+5.79\%)} \\
InstructRetro 48B \hfill  \emojiherb&  38.6 & 77.8 &  57.0 & 74.8 / 67.7 & 76.4 / 69.0 & 76.1 & 59.8 & 54.6 \\
\tiny{(w/ encoder, Avg: +6\%)} & \tiny{(+4.32\%)} & \tiny{(-0.38\%)} & \tiny{(+8.78\%)} & \tiny{(+5.80\%)} & \tiny{(+5.52\%)} & \tiny{(+6.43\%)} & \tiny{(+10.95\%)} & \tiny{(+5.41\%)} \\
\bottomrule
\end{tabular}
\end{table*}

\section{Experiments}
\label{sec:experiment}

In this section, we conduct comprehensive studies on the zero-shot capabilities of \emph{\method} and its GPT counterpart with RAG~(\instructgpt) across various downstream tasks to unveil the potential of Retro model after instruction tuning.


\subsection{Experimental setup}

\textbf{Datasets.}
To demonstrate the generalization of instruction tuning, we follow FLAN \citep{wei2021finetuned} and primarily focus on \textit{zero-shot evaluation} of {downstream tasks}.
Specifically, we consider two categories of open-ended QA tasks as well as text summarization tasks: (1) \textit{short-form QA or reading comprehension}, which expects short answers~(e.g., a few tokens) to be generated or extracted from the context, including Natural Question (NQ) \citep{kwiatkowski-etal-2019-natural}, TriviaQA \citep{joshi2017triviaqa}, NewsQA \citep{trischler2016newsqa}, SQuAD 1.1 \citep{rajpurkar2016squad}, SQuAD 2.0 \citep{rajpurkar2018suqad2}, Quoref \citep{dasigi2019quoref}, NarrativeQA \citep{kovcisky2018narrativeqa}, DROP \citep{dua2019drop}. 
To compare with baselines, we use the split from KILT benchmark~\citep{petroni-etal-2021-kilt} for NQ and TriviaQA. For the other tasks, we use the official splits; 
(2)~\textit{long-form QA}, which expects longer answer spans within a few sentences, including
doc2dial \citep{feng2020doc2dial}, two proprietary annotated car manual datasets~(people ask questions about the particular car models), and another proprietary annotated IT documentation dataset;
(3)~\textit{summarization}, which expects to summarize a long passage or context within a few sentences, including QMSum \citep{zhong2021qmsum}, SummScreenFd \citep{Chen2021SummScreenAD}, and GovReport \citep{Huang2021EfficientAF}.

\textbf{Retrieval-augmented generations~(RAG).} 
At pretraining, we use BERT embeddings to embed the retrieval database and support retrieval from trillions of tokens. 
For downstream task evaluation, we follow \retro \citep{borgeaud2022improving} and use task-specific corpus and state-of-the-art retrievers to retrieve the most relevant and high-quality information for the task. 
Specifically, 
for NQ, TriviaQA, doc2dial, and other long-form QA datasets, we use DRAGON+ \citep{lin2023train} as the retriever. 
We retrieve the top-$k=5$ nearest neighbors and concatenate them in the prompt. 
For the remaining QA and summarization tasks, we use the provided contexts in the datasets. 
An example of how we format the retrieved neighbors in the prompt is shown in Appendix Table \ref{table:squad_conv}.


\color{black}

\textbf{Models.}
\instructgpt 43B is our main baseline as it has the same decoder hyper-parameters as \method 48B  (e.g., number of transformer layers, hidden sizes, etc.) and was pretrained and instruction-tuned on the same datasets. 
We also compare them with \method~43B, which is the derivative of \method~48B by turning off the gate of cross-attention~(\textit{i.e.}, bypassing the encoder).
Note that we do instruction tuning on \retro without enabling its encoder. When we bypass the encoder during evaluation, \textbf{\method 43B solely serves as a GPT decoder} to align with the instruction tuning behaviors and simplify the inference.

For additional baselines, we also compare a wide range of state-of-the-art LLMs with comparable or larger sizes, including GPT-3 175B \citep{gpt3}, 
, GLaM 64B~\citep{du2021glam}, FLAN-LaMDA 137B~\citep{wei2021finetuned}, and
Llama 2 70B~\citep{llama2} with RAG\footnote{We evaluate both Llama 2 70B text model and instruction-tuned chat model with RAG, and report the best numbers.}.
Furthermore, we also compare \method with existing retrieval-augmented LLMs, including Retro 7.5B \citep{borgeaud2022improving}, Retro++ 9B \citep{wang2023shall}, Atlas 11B \citep{izacard2022atlas}, and Raven 11B \citep{huang2023raven}.

\begin{table}[t] 
    \centering
    \caption{\small Zero-shot evaluation on {four} long-form QA datasets. We use F1 as the evaluation metric. Car \#1 and \#2 are short for two annotated car manual datasets. The average \textit{relative} improvement of \method 43B across the long-form QA tasks is 10\% over \instructgpt 43B.}
    \label{tab:longqa}
    \vspace{5pt}
    \setlength{\tabcolsep}{3.25pt}
    \resizebox{0.95\linewidth}{!}
    {
    \begin{tabular}{l|cccc}
    \toprule
    & doc2dial & Car \#1 & Car  \#2 & IT Doc \\
    \midrule
    Llama 2 RAG \hfill 70B & 32.33 & {49.63} & 45.89 & 25.70 \\
    \instructgpt \hfill 43B & 32.87 & 58.18 & 50.88 & 31.40 \\
    \vspace{-1mm}
    InstructRetro \hfill 43B & {35.74} & \textbf{63.52} & \textbf{57.49} & \textbf{34.08}   \\
    \tiny{(w/o encoder, Avg: +10\%)}& \tiny{(+8.73\%)} & \tiny{(+9.18\%)} & \tiny{(+12.99\%)} & \tiny{(+8.54\%)} \\
    \vspace{-1mm}
    InstructRetro \hfill 48B & \textbf{35.95} & 63.16 & 56.82 & 34.07 \\
    \tiny{(w/ encoder, Avg: +10\%)}& \tiny{(+9.37\%)} & \tiny{(+8.56\%)} & \tiny{(+11.67\%)} & \tiny{(+8.50\%)} \\
    \bottomrule
    \end{tabular}
    }
\vspace{-1em}
\end{table}

\textbf{Other details.}
We use greedy decoding with the max output length to be 256. 
We truncate the generation when we encounter the special token \texttt{|<end-of-document>|} or role-switching from ``Assistant'' to ``User'' when completing the conversation.
All of the QA tasks are re-formatted in the conversational format. 
An example from the SQuAD~1.1 dataset in the conversational prompt format is shown in Appendix Table \ref{table:squad_conv}.

\subsection{Zero-shot evaluation on QA tasks}

We present the zero-shot evaluation results across eight short-form QA and reading comprehension datasets in Table \ref{tab:shortqa}. 
We also apply \method to four open-ended long-form QA datasets, as detailed in Table~\ref{tab:longqa}.
These datasets are representative of real-world applications, including  chatbots for IT support and customer service.  

\textbf{Instruction tuning post retrieval-augmented pretraining yields a better GPT decoder.}
From Table~\ref{tab:shortqa}, we observe that \method 43B shows consistent accuracy improvement upon its counterpart \instructgpt 43B across different datasets for short-form QA or reading comprehension tasks.
Notably, the average relative improvement of \method across all the short-form datasets is around 7\%.
Given that both \method 43B and \instructgpt 43B are pretrained and instruction tuned with identical datasets, hyper-parameters, and evaluation prompts, we attribute this consistent improvement to the training recipe of \method, which leverages continued pretraining with retrieval before instruction tuning. 
We hypothesize that retrieval-augmented pretraining enhances the capability of LLMs to utilize the information within the context (from both \retro encoder and decoder). The subsequent phase of instruction tuning further amplifies the effectiveness of \method in solving knowledge-intensive tasks.
To have a deeper understanding, we provide an ablation study in \S\ref{sec:ablation}.

From Table \ref{tab:shortqa}, we also show that \method 43B provides compelling performance than other state-of-the-art LLMs.
For example, \method 43B achieves better accuracy than Llama 2 with RAG on multiple tasks, close to FLAN-LaMDA 137B, which is 3$\times$ the size of \method 43B.

\textbf{Impact of Retro encoder for downstream tasks.}
We also notice that \method 48B and 43B perform very comparable from Table~\ref{tab:shortqa}.
We enable the \retro encoder for retrieval-augmented pretraining, while disabling the \retro encoder due to the lack of retrieved high-quality neighbors for instruction tuning. 
Note that we still perform retrieval-augmented generation for downstream tasks, where the retrieved contexts are put into the decoder of both \method 48B and 43B as part of the prompts.  
The only difference is whether we enable the cross attention gate in Figure \ref{fig:retro} to attend the \retro encoder in \method 48B or disable it in \method 43B.
When enabling the \retro encoder, we put the top-\emph{2} neighbors in the encoder to align with the pretraining behavior. 

This suggests that although \retro is proficiently trained to infer both with and without the neighbors in the encoder, it is more beneficial to align with the instruction tuning protocols and bypass the \retro encoder to solely serve as a GPT decoder during evaluation.
We think it is an important and promising future research direction to explore retrieval-augmented instruction tuning with the \retro encoder activated, especially when high-quality retrieval-augmented instruction tuning data is available.



\begin{figure*}[htp]
    \centering
    \begin{subfigure}{0.48\textwidth}
        \includegraphics[width=\linewidth]{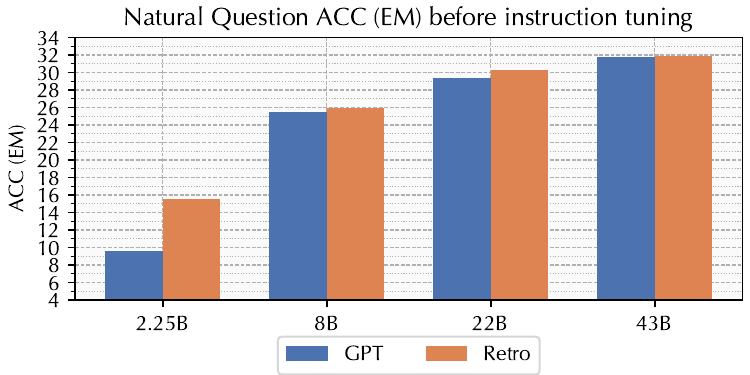}
        \caption{\small Before instruction tuning, the improvement of retrieval augmentation saturates when the size scales up.}
        \label{fig:nq}
    \end{subfigure}
    \hfill 
    \begin{subfigure}{0.48\textwidth}
        \includegraphics[width=\linewidth]{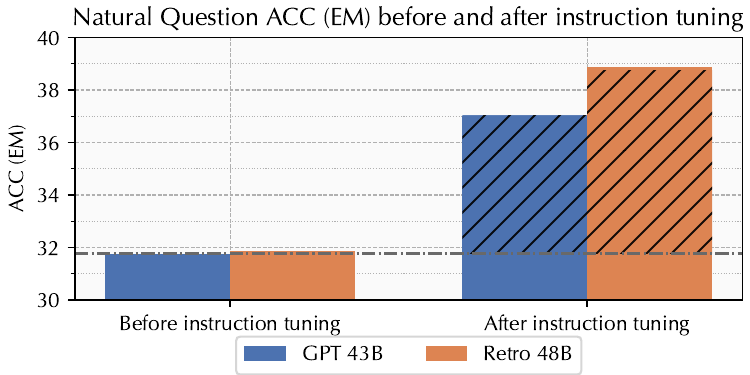}
        \caption{\small Instruction tuning further unveils the potential of retrieval augmentation even when the size scales up.}
        \label{fig:nq_inst}
    \end{subfigure}
    \vspace{-1em}
    \caption{Zero-shot accuracy (EM) of GPT and Retro before and after instruction tuning evaluated on the Natural Question dataset.}
    \label{fig:ablation-nq}
\end{figure*}

\textbf{\method demonstrates larger improvement on long-form QA datasets.}
When comparing the results of \method on short-form QA datasets and long-form QA datasets, we observe \method 43B demonstrates large relative accuracy improvements, achieving 10\% over the \instructgpt 43B.
As long-form QA tasks are generally more challenging than short-form QA tasks, such improvements further demonstrate the potential of retrieval-augmented pretraining. Again, the results of \method 43B and 48B results are very comparable, while \method 43B performs slightly better.

\begin{table}[t] 
 \centering
 \vspace{-.3cm}
    \caption{Zero-shot evaluation on three summarization datasets. We use the standard ROUGE scores as the evaluation metrics. The average \textit{relative} improvement of \method 43B across the long-form QA tasks is 16\% over \instructgpt 43B. }
    \label{tab:sum}
        \vspace{5pt}
        \setlength{\tabcolsep}{3.25pt}
    \resizebox{0.95\linewidth}{!}
        {
    \begin{tabular}{l|ccc}
    \toprule
     & \multicolumn{1}{l}{GovReport} & \multicolumn{1}{l}{SummFD} & \multicolumn{1}{l}{QMSum} \\ \midrule
     Llama 2 RAG \hfill 70B & 16.98 & 10.02 & 14.50 \\
    \instructgpt \hfill 43B & 12.59                         & 10.43                            & 15.06                     \\
    \vspace{-1mm}
    \method \hfill 43B & \textbf{17.46}                         & \textbf{10.93}                            & \textbf{15.61}                    \\ 
    \tiny{(w/o encoder, Avg: +16\%)}& \tiny{(+38.68\%)} & \tiny{(+4.79\%)} & \tiny{(+3.65\%)} \\
    \bottomrule
    \end{tabular}
    }
\end{table}

\subsection{Zero-shot evaluation on summarization tasks}
\label{app:summarization}
We also apply \method for summarization tasks, including QMSum \citep{zhong2021qmsum}, SummScreenFD \citep{Chen2021SummScreenAD}, and GovReport \citep{Huang2021EfficientAF}. Following the official metrics, we report the geometric mean of ROUGE scores (\textit{i.e.}, ROUGE1/2/L) for these summarization tasks. The zero-shot evaluation results are shown in Table \ref{tab:sum}.
From Table \ref{tab:sum}, we observe that \method consistently outperforms the \instructgpt on these summarization tasks, especially on the GovReport dataset with 4.87 ROUGE score improvement. Moreover, \method 43B consistently outperforms Llama 2 RAG 70B across three datasets. This experiment further confirms the generalizability of \method after instruction tuning and indicates that instruction tuning post retrieval-augmented pretraining yields a better GPT decoder.

\subsection{Ablation stuides}
\label{sec:ablation}

In this section, we conduct ablation studies to understand the source of improvements for \method. 
We show that both retrieval-augmented pretraining and instruction tuning are indispensable to unlock the potential of retrieval-augmented LLMs.

To understand how instruction tuning improves retrieval-augmented pretraining, we show the zero-shot accuracy~(Exact Match score) of  \retro and GPT on the Natural Question dataset before and after instruction tuning, as detailed in Figure \ref{fig:ablation-nq}. 
We observe that \retro achieves significantly better zero-shot accuracy than GPT when the number of parameters is relatively small (e.g., 2.25B). 
However, when scaling the size of parameters, the zero-shot performances of both GPT and Retro start to saturate.
We hypothesize that this saturation is mainly due to the poor instruction following capability of both pretrained foundation GPT and Retro models.
%

To remove the instruction-following bottleneck, we apply instruction tuning to further fine-tune both \retro 48B and GPT 43B. 
Instruction tuning largely mitigate the instruction following bottleneck for both GPT and Retro, resulting in a significant increase of their zero-shot performance on downstream tasks, respectively. 
Furthermore, once this bottleneck is alleviated, the benefits of retrieval augmentation at pretraining become more pronounced, as \method excels in leveraging and integrating evidence from retrieved context. Thus, we observe significant improvement of \method over \instructgpt again in Figure~\ref{fig:nq_inst}.
%
This ablation study confirms that our training recipe - both retrieval-augmented pretraining and instruction tuning are important for achieving high performance in  QA tasks. 

\subsection{Evaluation on MT-Bench}
We further evaluate \method and InstructGPT on the MT-Bench chat benchmark \citep{zheng2024judging} to access \retro's performance on general chat tasks. One may find more details in Appendix~\ref{app:mt-bench}.

\section{Conclusion}
\label{sec:conclusion}
In this paper, we introduce \method 48B, the largest LLM with retrieval-augmented pretraining and instruction tuning. 
Specifically, we start from a pretrained GPT model, and continue to pretrain the model with retrieval, which yields the retrieval-augmented foundation model \retro 48B.
After applying instruction tuning to \retro, \method 48B unveils the potential of retrieval-augmented pretraining and demonstrates significant zero-shot accuracy improvement over its GPT counterpart through our extensive experiments on a wide range of downstream tasks.
Moreover, our novel findings show that only using the GPT decoder backbone, {i.e.}, \method 43B, can achieve comparable accuracy, which sheds light on a promising direction to obtain a better GPT decoder through retrieval-augmented pretraining before instruction tuning.

\section*{Impact Statement}
Our \method, similar to the line of RAG studies, offers significant advancements in addressing the practical deployment and applications of LLMs, particularly in areas of factuality, downstream task accuracy, contextual understanding, and model efficiency.
Specifically, it significantly improves the generation of factual and grounded text with retrieval from high-quality databases, which helps mitigate misinformation and enhance public trust.
Additionally, InstructRetro boosts the accuracy of LLMs in various downstream tasks and demonstrates better capability in contextual understanding, which is important for reasoning tasks. 
Remarkably, it achieves comparable performance to LLMs two to three times its size, enhancing computational efficiency and environmental sustainability.
This positions \method as a practical tool in the ethical and safe deployment of LLMs.

\bibliography{main}
\bibliographystyle{icml2024}

\clearpage

\appendix

\onecolumn

\section{Details of Pretraining}
\label{app:pretraining}

\subsection{Pretraining corpus}
\label{app:corpus}
We prepared a pretraining dataset consisting of around 1.2 trillion tokens from English natural language data. 
Specifically, it consists of web-crawl data from Common Crawl, news data, conversational data, book data (e.g., Book3 and Book-Corpus2 from the Pile dataset \citep{gao2020pile}), scientific and multi-domain data (e.g., Wikipedia and the BigScience ROOTS corpus \citep{laurenccon2022bigscience}).

\subsection{Continued pretraining schedules}
\label{app:schedule}
Based on pretrained GPT models, we further pretrain \retro with retrieval augmentation on additional 100 billion tokens, which is around 25M samples with sequence length set to 4096. 
We list the pretraining hyper-parameter details of Retro-fitting in Table \ref{tab:pretrain_details}.  GPT-fitting uses the same training schedules as Retro-fitting.

All models use Adam optimizer \citep{adam} with $\beta_1=0.9$ and $\beta_2=0.95$. We employ the learning rate (LR) decay schedules with LR warmup samples of 16667 and LR decay samples of 23750000.

\begin{table}[tbh!]
\small
    \centering
    \caption{Detailed pretraining setup for standard pre-trained LMs and \method.}
    \label{tab:pretrain_details}
    {
    \begin{tabular}{l|cccccc}
    \toprule
\multicolumn{1}{l|}{\multirow{1}{*}{\textbf{Models Size}}}  &  LR & min LR & LR Decay Styles  & Batch Size & Pretraining Steps \\
\midrule
823M & 2e-5 & 2e-6 & cosine & 128  & 195.2k \\
2.25B & 2e-5 & 2e-6 & cosine & 256  & 97.6k \\
8.5B & 1e-5 & 1e-6 & cosine & 512  & 48.8K \\
22B & 1e-5 & 1e-6 & cosine & 512  & 48.8K \\
43B & 9e-6 & 9e-7 & cosine & 768  & 32.5k \\
\bottomrule
\end{tabular}
}
\end{table}

\color{black}
\subsection{Computational cost for continued pretraining}
\label{app:computational-cost}

We present the detailed computational cost of the continued pretraining step on additional 100B tokens for both \retro and GPT across different sizes in Table \ref{tab:pretrain_cost}. We can see that pretraining Retro brings around additional 35\% computational overhead than pretraining GPT, which mainly comes from the Retro encoder and cross-chunk attention to incorporate and fuse the retrieved neighbor information. Moreover, we can see that scaling up the size of Retro does not bring more computational overhead and remains around 35\%, shedding light on a promising way to retrieval-augmented pretraining.

A more useful perspective is looking at the overall pretraining cost. Since our Retro 48B starts from a pretrained GPT 43B on 1.1T tokens, it only need 2.58\% additional GPU hours in contrast to pretraining the GPT 43B on 1.2T tokens.
\begin{align}
  \frac{1.1\text{T} \times 1 + 0.1\text{T} \times (1 + 31\%)}{1.2\text{T} \times 1} = 102.58 \%
  \nonumber
\end{align}

\begin{table}[h]
\small
\centering
\caption{Pretraining cost of the continued pretraining on 100B tokens for \retro and GPT across different sizes.}
\label{tab:pretrain_cost}
\begin{tabular}{l|c||c|c||c|c}
\toprule
GPT      & training on 100B token & Retro  & training on 100B token & \shortstack{Additional Overhead\\(on 100B tokens)} & \shortstack{Additional Overall Overhead\\(on 1.2T tokens)}  \\
\midrule
823M & 1408 GPU Hours  &  878M & 1920 GPU Hours      & 36\%        &  3.00\%      \\
2.25B   & 3226 GPU Hours  & 2.5B  & 4096 GPU Hours      & 27\%     &  2.25\%        \\
8.5B  & 12698 GPU Hours  & 9.5B & 17325 GPU Hours     & 37\%       &  3.08\%       \\
22B  & 37888 GPU Hours  & 24B & 52152 GPU Hours     & 37\%         &  3.08\%     \\
43B  & 53329 GPU Hours  & 48B & 69995 GPU Hours     & 31\%         &  2.58\%     \\
\bottomrule
\end{tabular}
\end{table}

\color{black}
\section{Details of retrieval database}
\label{app:faiss}
\paragraph{Retrieval Database.} We use the whole pretraining corpus as our retrieval database, consisting of 1.2 trillion tokens as mentioned in Appendix \S\ref{app:corpus}. 
Our pretraining dataset with 1.2 trillion tokens yields a retrieval database consisting of 19B chunks in total with chunk size $m=64$. To support fast similarity searches with billions of chunks, we implement the database index with Faiss index~\citep{faiss}.
Given the BERT embeddings of an input chunk $C_i$, Faiss can return the approximate $k$ nearest neighbor of $C_i$ within a few milliseconds.

\subsection{Faiss index configuration}
We use the Faiss index \citep{faiss} as the implementation for the dense retriever to search for approximate nearest neighbors in the BERT embedding space.
We configure the Faiss index as follows:
\begin{itemize}[leftmargin=.7em,topsep=0pt,itemsep=0.0pt]
\item \textbf{Preprocessing}: We use Optimized Product Quantization \citep{opq} to apply a rotation to the input vectors to make them more amenable to PQ coding \citep{pq}.
\item \textbf{Indexer}: We use Inverted File Index (IVF) with $2^{22}$ centroids and accelerate it with Hierarchical Navigable Small World (HNSW) graphs \citep{hnsw}.
\item \textbf{Encoding}: We adopt PQ encoding that compresses the dense embedding vector into 64 bits.
\end{itemize}
As a result, we can achieve \textit{4ms} per query over the whole pretraining corpus via batch queries averaged for each chunk with less than 1TB memory usage as our max throughput. 
Given a single query, the latency of the response is around $0.1s$ per query.
We also note that increasing the number of $K$ in the query does not yield slower query speed.
During pretraining, we follow \citet{borgeaud2022improving} to pre-compute the nearest neighbors and save the data for pretraining.

\color{black}
\subsection{Computational cost on building retrieval database}

Building a Faiss index involves several steps. We detail each step with its associated computational cost as below:
\begin{itemize}[leftmargin=.7em,topsep=0pt,itemsep=0.0pt]
\item \textbf{Embedding the retrieval database into dense BERT embeddings.} Given the chunk size of $m=64$ tokens, we embed every chunk of text corpus with BERT-large-cased. The computational cost to embed the text corpus is around 6.22M chunks per GPU hour given one A100 GPU. For our 19B chunk database, it takes around 3054 GPU hours in total. 
\item \textbf{Train the Faiss index.} This involves determining a smaller number of centroids to cluster the whole corpus embeddings and initializing the HNSW graph. The computational cost of training the Faiss index depends on the number of corpus embeddings and the number of centroids. Given our setup, we train the faiss index based on 600M chunks uniformly sampled from the whole retrieval database. The computational cost of this step is less than 4 hours with one DGX A100 node. 
\item \textbf{Add the embedded corpus to the Faiss index.} After the index has been trained, the index centroids and HNSW graph are determined, but the index itself is still empty. In this step, we add the whole dense corpus embeddings to the index data structure. 
The computational cost of adding the corpus to the index is around 192 CPU hours within one DGX A100 node. Moreover, it can be purely done within a CPU node to save computational cost.
\item \textbf{Query the Faiss index.} As mentioned above, we can achieve \textit{4ms} per query over the whole pretraining corpus via batch queries averaged for each chunk with less than 1TB memory usage as our max throughput. The computational cost to query over 100B tokens in our continued pretraining step is around 1736 CPU hours within a DGX A100 node. Moreover, this step can also be purely done within a CPU node to save computational cost and can run in parallel to further speed up the querying.
\end{itemize}

In summary, the overall computational cost of building Faiss index is marginal compared to the pretraining cost, especially considering the benefits of retrieval-augmentation pretraining, which further unlocks the potential of instruction tuning. Thus we believe that it is a promising direction to pretrain with retrieval augmentation.

\subsection{Ablation studies on Faiss index confirations}

\paragraph{Faiss training-time configuration.}

We conduct ablation studies on the quantization techniques using two index configurations on two datasets: the whole pretraining dataset and the Wikipedia Corpus. We highlight the configuration setup in Table \ref{tab:faiss-pq} below.

\begin{table}[h!]
\color{black}
\caption{Ablation studies on Faiss product quantization (PQ) on two different retrieval databases.}
\label{tab:faiss-pq}
\centering
\begin{tabular}{ll|cc}
\toprule
\multicolumn{2}{c}{{}}                                                                                                       & \shortstack{ Retrieval Index for\\ Full   Pretraining Corpus} & \shortstack{ Retrieval Index for\\ Wikipedia   Corpus} \\ \midrule

\multicolumn{2}{c|}{{ \#/ chunks}}                                                                                                                                                                 & { 19B}                                 & { 66M}                             \\
\midrule

{ }                                                                                                     & { Dimension Reduction}                                            & { OPQ64\_128}                                    & { No Reduction}                           \\

{ }                                                                                                     & \multicolumn{1}{l|}{{ Approximate Search}} & { IVF4194304\_HNSW32}                            & { IVF262144\_HNSW32}                      \\
\multirow{-3}{*}{{ Configuration}}                                                                      & \multicolumn{1}{l|}{{ Encoding}}           & { PQ64}                                          & { Flat Encoding}                          \\
\midrule
{ }                                                                                                     & { K=2}                                                            & { 0.004 s/query}                                 & { 0.01 s/query}                           \\

{ }                                                                                                     & { K=20}                                                           & { 0.004 s/query}                                 & { 0.01 s/query}                           \\

{ }                                                                                                     & { K=200}                                                          & {0.0045 s/query}                                & { 0.01 s/query}                           \\
\multirow{-4}{*}{Query Speed} & { K=2000}                                                                                & 0.004 s/query                                                        & 0.01 s/query                                                  \\
\bottomrule
\end{tabular}
\end{table}

Following the official guide of Faiss\footnote{\url{https://github.com/facebookresearch/faiss/wiki/Guidelines-to-choose-an-index}}, we initialize two Faiss indexes based on the sizes of two retrieval databases: the full pretraining corpus with 19B chunks and the Wikipedia corpus with 66M chunks. 
We applied product quantization \citep{opq,pq} to the full pretraining corpus to reduce the dimensionality and save the index memory to support loading the full pretraining corpus, while applying uncompressed flat encoding to the Wikipedia corpus as a comparison. 
We benchmark the querying speed for a batch of 40K dense embeddings and evaluate the query speed for two indexes. 

From Table \ref{tab:faiss-pq}, we can see that applying product quantization can not only help compress the index and save memory usage but also help improve the query speed, which is critical when scaling up the retrieval database. We can also see that increasing the number of $K$ for $K$ nearest neighbor searchers barely impacts the query speed.

\color{black}
\paragraph{Faiss query-time configuration.}

For our index configuration with interveted file index structures and HNSW graph, the hyper-parameter \texttt{nprobe} and \texttt{efSearch} play important roles in the query time of Faiss, as detailed in Table \ref{tab:faiss-param}.

\begin{table}[h!]\small
\color{black}
\caption{Important querying-time hyper-parameters for our Faiss index.}
\label{tab:faiss-param}
\begin{tabular}{@{}llll@{}}
\toprule
index type   & Index class & runtime parameter & comments                                                    \\ \midrule
IVF*, IMI2x* & IndexIVF*   & nprobe            & the main parameter to adjust the   speed-precision tradeoff \\
HNSW*        & IndexHNSW   & efSearch          & the depth of the HNSW search                                \\ \bottomrule
\end{tabular}
\end{table}

To select a proper set of query-time hyper-parameters with a good tradeoff of recall and speed, we conduct ablation studies with varying \texttt{nprobe} and \texttt{efSearch}. Specifically, we use the retrieval index built on the whole pretraining corpus, query the index with randomly sampled 10K chunks from the pretraining corpus, and evaluate the recall accuracy of retrieving the query chunk itself given top-$K=2000$. The query time and corresponding recall accuracy with different hyper-parameters are shown in Table \ref{tab:faiss-param-res}. 
We thus select \texttt{nprobe}=4096 and \texttt{efSearch}=32 in our retrieval-augmented pretraining, considering its fast querying time without sacrificing the recall accuracy significantly.

\begin{table}[h!]
\centering
\caption{Trade-off between speed and recall accuracy given different query-time hyper-parameters.}
\label{tab:faiss-param-res}
\color{black}
\begin{tabular}{@{}lll@{}}
\toprule
                           & Time (per query) & Acc (K=2000) \\
\midrule
efSearch=256, nprobe=65536 & 21ms             & 1.00         \\
efSearch=64, nprobe=16384  & 6ms              & 0.96         \\
\textbf{efSearch=32, nprobe=4096} & \textbf{4ms}              & \textbf{0.93}         \\
efSearch=16, nprobe=4096   & 2.5ms            & 0.88        \\
\bottomrule
\end{tabular}
\end{table}

\clearpage

\color{black}
\section{Qualitative examples}
\subsection{An example From the instruction tuning data}
\label{app:example}

Table \ref{table:conversation} below is a data example from the instruction tuning data. 

\begin{table}[h!]
    \centering
    \caption{A sample conversation from our instruction tuning dataset. The black part serves as the context, while the black part is masked as the answer mask, on which the LM loss is computed.}
    \label{table:conversation}
    \begin{tabular}{m{2cm}|m{10cm}}
    \hline
    \textbf{Speaker} & \vspace{3mm}\textbf{Conversation}\vspace{3mm} \\
    \hline
    System & \vspace{3mm} This is a chat between a user and an artificial intelligence assistant. The assistant gives helpful, detailed, and polite answers to the user's questions based on the context. The assistant should also indicate when the answer cannot be found in the context. \vspace{3mm} \\
    \hline
    User & \vspace{3mm}My bicycle tire has a puncture and while I have a pump with me, I don't have a patch repair kit. Is there a way to keep the tire from running out of air until I can get to a bicycle shop?\vspace{3mm} \\
    \hline
    Assistant & \vspace{3mm}Yes, you could use duct tape to seal the puncture.\vspace{3mm} \\
    \hline
    User &     \vspace{3mm} Can you give me a tutorial on how to use the duct tape to seal the puncture?     \vspace{3mm} \\
    \hline
    {\color{black} Assistant} & \vspace{3mm}{\color{black} Sure, here's a tutorial on how to use duct tape to seal a puncture in a tire:\newline 
    1. Start by cleaning the area around the puncture with a damp cloth.\newline
    2. Cut a piece of duct tape large enough to cover the puncture, plus a few millimeters of extra.\newline
    3. Remove the backing from the duct tape, and press it firmly over the puncture.\newline
    4. Rub the tape firmly with your fingers to ensure a good seal.\newline
    5. You can also use a roller or a flat item like a ruler to make sure that the tape is pressed firmly against the puncture.\newline
    6. Trim away any excess tape, and your puncture should now be sealed. }\vspace{3mm}\\
    \hline
    \end{tabular}
\end{table}

\clearpage

\subsection{An example From the downstream QA dataset: SQuAD 1.1}
We re-format all of the QA tasks in the conversational format. 
An example from the SQuAD 1.1 dataset with the conversational format is shown in the Table \ref{table:squad_conv} below.

\begin{table}[h!]
    \centering
    \caption{A sample conversation prompt template from the SQuAD 1.1 dataset.}
    \label{table:squad_conv}
    \begin{tabular}{m{2cm}|m{10cm}}
    \toprule
    \textbf{Speaker} & \textbf{Conversation} \\
    \midrule
    System & System: This is a chat between a user and an artificial intelligence assistant. The assistant gives helpful, detailed, and polite answers to the user's questions based on the context. The assistant should also indicate when the answer cannot be found in the context. \\ \midrule

    User & title: , source: The pound-force has a metric counterpart, less commonly used than the newton: the kilogram-force (kgf) (sometimes kilopond), is the force exerted by standard gravity on one kilogram of mass. The kilogram-force leads to an alternate, but rarely used unit of mass: the metric slug (sometimes mug or hyl) is that mass that accelerates at 1 ms-2 when subjected to a force of 1 kgf. The kilogram-force is not a part of the modern SI system, and is generally deprecated; however it still sees use for some purposes as expressing aircraft weight, jet thrust, bicycle spoke tension, torque wrench settings and engine output torque. Other arcane units of force include the sthène, which is equivalent to 1000 N, and the kip, which is equivalent to 1000 lbf. \newline

    Based on the above article, answer a question. What is the seldom used force unit equal to one thousand newtons? \\
    \midrule
    Assistant & The answer is \\
    \bottomrule
    \end{tabular}
\end{table}

\clearpage
\section{Experimental results on MT Bench}
\label{app:mt-bench}


We evaluate \method and InstructGPT on the MTBench chat benchmark \citep{zheng2024judging} to understand \retro performance on general chat tasks. 
We use the Tulu-v2 \citep{wang2023far} alignment dataset as the instruction tuning dataset to enhance the chat capabilities for both pretrained GPT model and \retro model.
The detailed breakdown of the MT-Bench result is shown in Table \ref{tab:mt_bench_comparison}.

\begin{table}[ht]
\centering
\caption{Performance comparison of MT-Bench models}
\label{tab:mt_bench_comparison}
\begin{tabular}{lcc}
\toprule
MT-Bench & \method-Tulu-v2-43B & InstructGPT-Tulu-v2-43B \\ 
\midrule
Writing    & \textbf{8.85} & 8.15 \\
Roleplay   & 7.75 & \textbf{7.80} \\
Reasoning  & \textbf{5.40} & 4.75 \\
Math       & \textbf{3.15} & 2.35 \\
Coding     & {3.40} & \textbf{4.10} \\
Extraction & \textbf{6.80} & 6.75 \\
STEM       & \textbf{8.58} & 8.53 \\
Humanities & \textbf{9.68} & 9.10 \\
\midrule
\addlinespace
Turn 1     & \textbf{6.89} & 6.67 \\
Turn 2     & \textbf{6.51} & 6.21 \\
Avg        & \textbf{6.70} & 6.44 \\
\bottomrule
\end{tabular}
\end{table}

In Table \ref{tab:mt_bench_comparison}, we show that on average \method outperforms InstructGPT across different turns.
We also observe that Retro performs slightly lower than GPT in domains such as role play and coding. We think that the main reason is that we do not cover coding and role-playing related datasets in the pretraining dataset of the base GPT model, and thus retrieval-augmented pretraining could make little difference.



\color{black}

\section{Potential Negative Social Impacts}
In this section, we discuss a few potential negative social
impacts shared by the current line of LLM research. 
First, similar to other very capable LLMs, \method can generate non-factual but persuasive text across a wide range of topics. This ability can be maliciously exploited to create and spread disinformation or misinformation at scale.
Second, \method is trained on vast datasets collected from the internet, which may include personal or private information.
In addition, the retrieval database used in \method may contain private information as well.
Third, although Retro framework was found to be effective reduce toxic generations~\cite{wang2023shall}, it still reflect and can amplify the biases present in the training data as other LLMs.

\end{document}